\documentclass[runningheads,envcountsect]{llncs}

\newif\iftightsubmission
\tightsubmissionfalse
\newif\ifcameraready
\camerareadytrue

\usepackage[T1]{fontenc}
\usepackage[protrusion=true,expansion=false]{microtype}
\usepackage{graphicx}
\usepackage{subcaption}
\usepackage{wrapfig}
\usepackage{booktabs}
\usepackage{amsmath}
\usepackage{amssymb}
\usepackage{mathtools}
\usepackage{multirow}
\usepackage{algorithm}
\usepackage{algorithmic}
\usepackage{marvosym}
\usepackage{hyperref}
\usepackage[capitalize,noabbrev]{cleveref}
\usepackage[textsize=tiny]{todonotes}

\begin{document}

\title{Spectral Overfitting in Noisy Linear Probing of Pretrained Representations}
\titlerunning{Spectral Overfitting in Noisy Linear Probes}

\ifcameraready
\author{Zice Wang\inst{1,2}\textsuperscript{\Letter} \and
Zhenyu Zhang\inst{3}\textsuperscript{\Letter}}
\authorrunning{Z. Wang and Z. Zhang}
\institute{
School of Future Technology, Northeastern University, Shenyang, China\\
\email{wangzc1@mails.neu.edu.cn}
\and
School of Computer Science and Engineering, Northeastern University, Shenyang, China
\and
School of Computer Science and Technology, Beijing Institute of Technology, Beijing, China\\
\email{charlie@bit.edu.cn}
}
\else
\authorrunning{}
\fi

\maketitle

\begin{abstract}
Frozen pretrained features are often treated as a safe interface for downstream learning: only a small linear readout is trained, while the backbone is fixed. We show that this readout can still overfit noisy labels in a structured way. A label-blind PCA rank sweep reveals a sharp spectral pattern: under label noise, exposing all pretrained directions can hurt clean accuracy, and intermediate ranks often recover much of the lost performance. Rank-matched random projections help less, and measured between-class signal is strongly concentrated in leading PCs. The pattern appears across three ImageNet-pretrained backbones on CIFAR-10, with gains up to $36.0\pm0.8$ points over the default full-rank probe at 40\% noise. Tuned full-rank probes outperform validation-selected PCA probes, so we present the sweep as a diagnostic of spectral overfitting rather than a competitive noisy-label method.
\end{abstract}

\section{Introduction}
\label{sec:intro}

Transfer learning with frozen pretrained backbones is widely used in practice: extract ImageNet-pretrained features and train a linear probe \cite{kornblith2019similarity}. Freezing the backbone removes representation optimization from the downstream problem, but not readout overfitting. With noisy downstream labels \cite{zhang2021understanding}, exposing every feature dimension can add more capacity than useful signal.

This raises a diagnostic question: where in the pretrained feature spectrum does a noisy linear probe overfit? We test whether the spectrum is ordered enough that exposing different numbers of principal components reveals a systematic transition from useful signal to noise-fitting capacity, and whether the effect depends on backbone capacity, task difficulty, or noise rate.

We study \textbf{noisy linear probing on fixed pretrained features}. Given features $H \in \mathbb{R}^{N \times D}$, we fit PCA on training features and train probes on the top $d$ components. The backbone never sees noisy labels, so changing $d$ only changes the readout. This gives a controlled way to vary readout capacity along unsupervised spectral directions. Rank-matched random projections separate spectral ordering from dimension reduction.

PCA is a classical compression and denoising tool \cite{jolliffe1986principal}, but the noise here is not feature noise: PCA is label-blind, and corruption enters only through probe labels. Full-rank PCA would merely reparameterize a linear probe; the informative perturbation is the rank restriction it enables. We ask whether removing lower-variance directions reveals that noisy probes were using them to fit corrupted labels. Clean-label sweeps and rank-matched random projections test whether the effect is more than discarding useless dimensions.

Across ResNet-18, VGG-16, and EfficientNet-B0 on CIFAR-10, intermediate PCA ranks outperform the default full-rank probe under label noise, with gains up to $36.0\pm0.8$ points at 40\% noise. Random projections help less than PCA, while direct measurements show that leading PCs preserve nearly all between-class signal. Clean-validation controls reproduce the rank-sweep pattern, but tuned full-rank probes outperform the selected PCA probes in every tested setting. We therefore treat the sweep as a diagnostic, not a replacement for probe regularization. CIFAR-100 needs stronger noise for a clear elbow; one noisy fine-tuning control does not reproduce the fixed-feature pattern.

Our contributions are a controlled rank-sweep diagnostic for noisy fixed-feature probing; cross-backbone measurements connecting concentrated class signal, rank, and noisy-probe overfitting; and random-projection, regularization, and boundary controls that delimit what the diagnostic does and does not establish. PCA truncation itself is classical; the contribution is the empirical characterization of its rank path under noisy linear probing.

\begin{figure*}[t]
	\centering
	\includegraphics[width=0.92\linewidth]{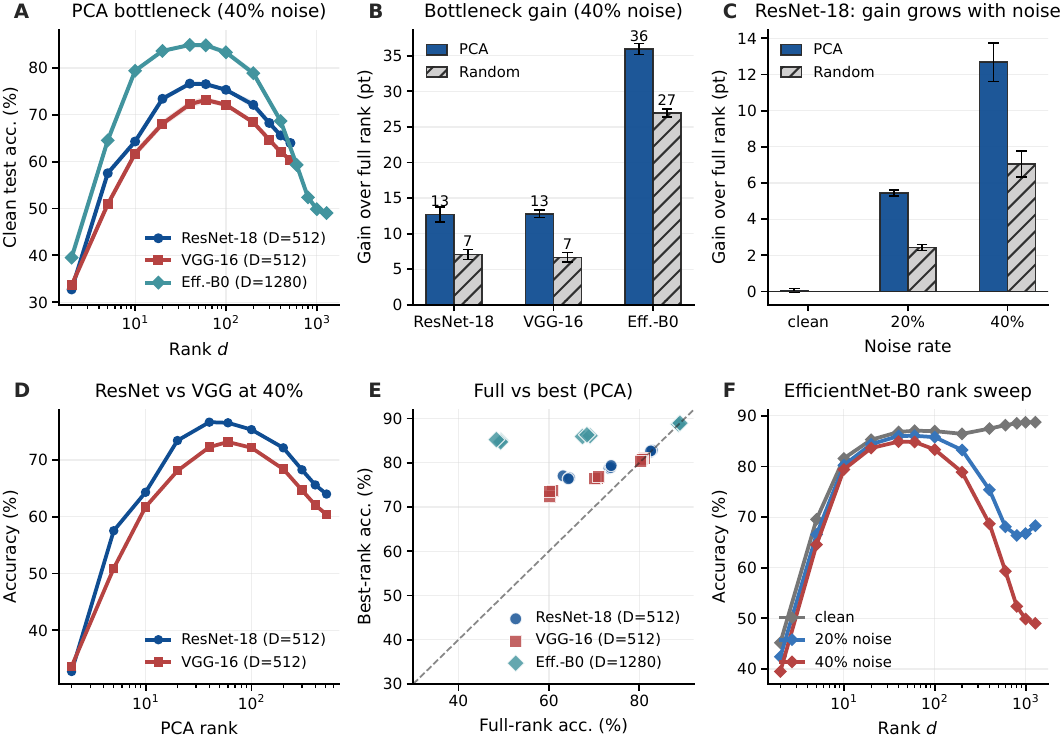}
	\caption{\textbf{Full-rank features can overfit noisy probes.} Frozen ImageNet-pretrained backbones on CIFAR-10 (5K train subset). (A) PCA rank sweep at 40\% noise. (B) PCA vs.~random gain. (C) Gain vs.~noise (ResNet-18, 3 seeds). (D) ResNet vs.~VGG. (E) Full-rank vs.~best-rank. (F) EfficientNet-B0 sweep.}
	\label{fig:fixed_probe_main}
\end{figure*}

\section{Related Work}
\label{sec:related}

Linear probing is widely used to evaluate pretrained representations \cite{kornblith2019similarity}. Label noise is known to encourage memorization and degrade generalization \cite{zhang2021understanding,ma2018dimensionality}; in this work, the noise affects only the linear readout. Regularized full-rank linear probes, including $L_2$-penalized logistic regression, are standard deployment baselines for this setting. Implicit-bias analyses \cite{soudry2018implicit,gunasekar2018characterizing} and spectral-bias work \cite{rahaman2019spectral} address training dynamics. Our use of ``spectral'' instead refers to fixed feature covariance and the number of PC directions exposed to a noisy probe. PCA is a classical tool for low-rank approximation, compression, and denoising of feature observations \cite{jolliffe1986principal}. The Information Bottleneck \cite{tishby2015deep} and intrinsic-dimension estimates \cite{ansuini2019intrinsic} also motivate compression. We use PCA rank as an explicit label-blind diagnostic restriction rather than as a supervised correction or a direct feature-noise denoising method.

\section{Rank-Limited Linear Probing}
\label{sec:method}

\subsection{Problem Setting}

Let $\phi: \mathcal{X} \to \mathbb{R}^D$ be a frozen feature extractor---the penultimate layer of an ImageNet-pretrained network. For a downstream training set $\{(x_i, y_i)\}_{i=1}^N \subset \mathcal{X} \times [K]$, we first extract raw features $\tilde{\mathbf h}_i=\phi(x_i)$ and explicitly center them as $\mathbf h_i=\tilde{\mathbf h}_i-N^{-1}\sum_j\tilde{\mathbf h}_j$. Thus $\mathbf{H} \in \mathbb{R}^{N \times D}$ satisfies $\sum_i \mathbf{h}_i = 0$ by construction. We then corrupt a fraction $\eta$ of training labels with symmetric label noise, producing $\tilde{y}_i$, and train only a linear probe. The backbone never sees $\tilde{y}_i$.

Let $\mathbf{\Sigma} = \frac{1}{N}\mathbf{H}^\top\mathbf{H}$ be the empirical covariance, with eigendecomposition $\mathbf{\Sigma} = \mathbf{V}\mathbf{\Lambda}\mathbf{V}^\top$, where $\mathbf{V}=[\mathbf{v}_1,\dots,\mathbf{v}_D]$ and $\lambda_1 \ge \dots \ge \lambda_D \ge 0$. For any rank $d \le D$, define the projected features $\mathbf{Z}_d = \mathbf{H}\,\mathbf{V}_{1:d} \in \mathbb{R}^{N\times d}$, where $\mathbf{V}_{1:d}$ contains the first $d$ principal directions. A multinomial logistic regression probe $f_d(\mathbf{z}) = \text{softmax}(\mathbf{W}\mathbf{z} + \mathbf{b})$, with $\mathbf{W}\in\mathbb{R}^{K\times d}$, is trained on $\{(\mathbf{z}_i, \tilde{y}_i)\}$ and evaluated on clean test labels.

\subsection{Bias--Variance Decomposition}

The effect of $d$ on clean test risk can be viewed through a bias--variance lens. Let $\mathcal{E}(d)$ denote the expected clean test error of the probe trained on $\mathbf{Z}_d$ with noisy labels:

\begin{equation}
    \mathcal{E}(d) = \underbrace{\mathcal{B}(d)}_{\text{lost signal}} \;+\; \underbrace{\mathcal{V}(d,\eta)}_{\text{noise-fitting}},
    \label{eq:bias_variance}
\end{equation}

where $\mathcal{B}(d)$ is approximation error from discarding semantic directions, and $\mathcal{V}(d,\eta)$ is excess variance from fitting corrupted labels. The signal term decreases with $d$; the noise term increases with $d$ and $\eta$.

Pretrained vision features often have concentrated spectra and class-structured directions \cite{papyan2020prevalence,ansuini2019intrinsic}. In our experiments, early PCs capture most CIFAR-10 feature variance. Later components can contain useful signal, but they also increase the number of directions available to fit corrupted labels. The rank sweep measures the resulting trade-off.

This interpretation differs from applying PCA as a generic denoiser. PCA is computed once from unlabeled feature vectors and does not observe which labels are corrupted. Its only effect is to choose a nested sequence of readout subspaces. A beneficial rank therefore means that the discarded directions were more useful for fitting noisy labels than for improving clean test accuracy under the given probe and dataset. Conversely, when labels are clean or when discarded directions contain task-relevant class information, full rank can be optimal. This is why we report clean-label sweeps and rank-matched random projections rather than treating PCA improvement as an automatic consequence of low-rank compression.

\subsection{Rank Sweep Protocol}

For each backbone, dataset, and seed we:

\begin{enumerate}
    \item Extract $\mathbf{H} \in \mathbb{R}^{N\times D}$ from the frozen backbone.
    \item Fit PCA on centered training features $\mathbf{H}$.
    \item Corrupt a fraction $\eta$ of training labels (symmetric noise).
    \item For each $d$ in a logarithmic rank grid, project features to $d$ dimensions via $\mathbf{V}_{1:d}$, train a multinomial logistic regression ($L_2$ penalty $C=1$), and evaluate clean test accuracy.
\end{enumerate}

The full-rank probe ($d=D$) with $C=1$ serves as the default baseline. We report the best observed rank to characterize the effect size of the spectral restriction, and we also run a clean-validation selection protocol: split the 5K training subset into 4K fit and 1K validation examples, corrupt only the fit labels, choose $d$ by validation accuracy, and report clean test accuracy for the selected rank. This separates the retrospective rank-sweep diagnostic from a deployable selection rule when a small clean validation set is available.

\subsection{Random Projection Baseline}

To separate spectral ordering from generic capacity reduction, we measure isotropic random orthogonal projections. For each rank $d$, we sample $\mathbf{Q} \in \mathbb{R}^{D\times d}$ from a standard normal distribution, orthonormalize via QR decomposition, and train the same logistic regression on $\mathbf{H}\mathbf{Q}$. This is related to Johnson--Lindenstrauss embeddings \cite{william1984extensions}: it reduces dimension without using the empirical spectrum. The PCA--random gap estimates the value of the ordered basis beyond rank alone.

\subsection{Conditions for Beneficial Truncation}
\label{sec:when_helps}

The rank sweep traces a bias--variance trade-off. Let the probe trained on $d$ principal components have clean test error $\mathcal{E}(d) = \mathcal{B}(d) + \mathcal{V}(d,\eta)$, where $\mathcal{B}(d)$ is the signal lost by discarding components beyond $d$, and $\mathcal{V}(d,\eta)$ is excess variance from fitting label noise $\eta$ in the retained dimensions. The full-rank probe ($d=D$) minimizes $\mathcal{B}$ but maximizes $\mathcal{V}$; truncation helps when $\mathcal{V}$ grows faster with $d$ than $\mathcal{B}$ shrinks.

Let $\lambda_1 \ge \dots \ge \lambda_D$ be the eigenvalues of $\mathbf{\Sigma}$, and let $s_k = \frac{1}{K}\sum_{c=1}^K \|\boldsymbol{\mu}_c^\top\mathbf{v}_k\|^2$ denote the between-class signal along component $k$, where $\boldsymbol{\mu}_c$ is the centroid of class $c$ in feature space. The tail sum $S_{\mathrm{tail}}(d)=\sum_{k>d}s_k$ is a proxy for signal discarded by truncation. On the variance side, standard complexity bounds for linear predictors grow with the effective dimension, all else equal \cite{kakade2008complexity}. Symmetric label noise makes this extra capacity harmful because some training labels are uncorrelated with the clean decision rule.

As qualitative bookkeeping, beneficial truncation requires avoided noise-fitting to exceed discarded signal,
\begin{equation}
    a\,S_{\mathrm{tail}}(d) \;<\; b\,\eta\,(D-d),
    \label{eq:gain}
\end{equation}
where $D-d$ counts removed readout directions and $a,b>0$ collect problem-dependent scale, regularization, and geometry. This inequality is an interpretive criterion, not a theorem or an estimator of the optimal rank: without assumptions fixing $a$ and $b$, it only identifies quantities that the experiments must measure. It highlights two conditions:

\begin{enumerate}
    \item \textbf{Spectral concentration:} The leading $k^* \ll D$ components capture most downstream signal, i.e.\ $\sum_{k=1}^{k^*} s_k \approx \sum_{k=1}^D s_k$. Then for $d \ge k^*$, the discarded-signal proxy is small.
    \item \textbf{Sufficient noise:} The label noise rate $\eta$ is large enough that reducing probe capacity offsets the residual signal loss. At low $\eta$, full rank can remain optimal.
\end{enumerate}

This view organizes, rather than proves, the three empirical regimes. On CIFAR-10, measured spectra and between-class signal are concentrated and 20--40\% noise is enough for intermediate ranks to help. On CIFAR-100, 60--80\% noise is needed for a clear gain. Random projections mix high- and low-variance directions instead of following the empirical spectrum, which is consistent with their smaller gain but does not by itself establish a causal mechanism.

The rank sweep measures whether these conditions hold for a given pretrained backbone and downstream task.

\section{Experimental Setup}
\label{sec:setup}

\paragraph{Datasets.}
We use CIFAR-10 and CIFAR-100 \cite{krizhevsky2009learning}. For the main fixed-feature experiments we use a 5,000-example random training subset, matching the low-data probing regime where label noise is most consequential. Test sets remain complete (10,000 examples).

\paragraph{Noise.}
Symmetric label noise: for a rate $\eta$, each selected training label is replaced uniformly by an incorrect class. Test labels are never corrupted.

\paragraph{Backbones.}
We evaluate three ImageNet-pretrained architectures from \texttt{timm} using the library default pretrained weights:
\begin{itemize}
    \item \texttt{resnet18} ($D=512$ penultimate features) \cite{he2016deep}
    \item \texttt{vgg16} ($D=512$ features)
    \item \texttt{efficientnet\_b0} ($D=1280$ features) \cite{tan2019efficientnet}
\end{itemize}
All images are resized to $224\times224$ and normalized with ImageNet statistics. We use \texttt{forward\_features}; spatial feature maps are globally averaged, and classifier heads are removed. No backbone parameter is updated.

\paragraph{Rank grid.}
$d\in\{2,5,10,20,40,60,100,200,300,400,512\}$ for $D=512$; extended to 600,800,1000,1280 for EfficientNet-B0.

\paragraph{Probe.}
Multinomial logistic regression with $L_2$ penalty uses inverse regularization strength $C=1$ unless otherwise stated, the LBFGS solver, and at most 1000 iterations. We center training features before PCA and use full SVD; test features use the training mean and PCs. Seeds $\{0,1,2\}$ control both the NumPy-random 5K subset indices and label corruption, making the indices exactly reproducible from the released code. Tuned controls evaluate $C\in\{10^{-3},3\cdot10^{-3},10^{-2},3\cdot10^{-2},10^{-1},3\cdot10^{-1},1,3,10,30,100\}$. Experiments use \texttt{timm} 1.0.27, scikit-learn 1.9.0, NumPy 2.4.6, and SciPy 1.17.1.

\paragraph{Random projection.}
For each rank $d$, we sample $\mathbf{Q}\sim\mathcal{N}(0,1)^{D\times d}$, orthonormalize via QR, and train the same probe on $\mathbf{H}\mathbf{Q}$.

\section{Results}
\label{sec:results}

\subsection{Pretrained Feature Spectra Concentrate Downstream Signal}
\label{sec:spectral_analysis}

\begin{figure}[t]
\centering
\includegraphics[width=0.88\linewidth]{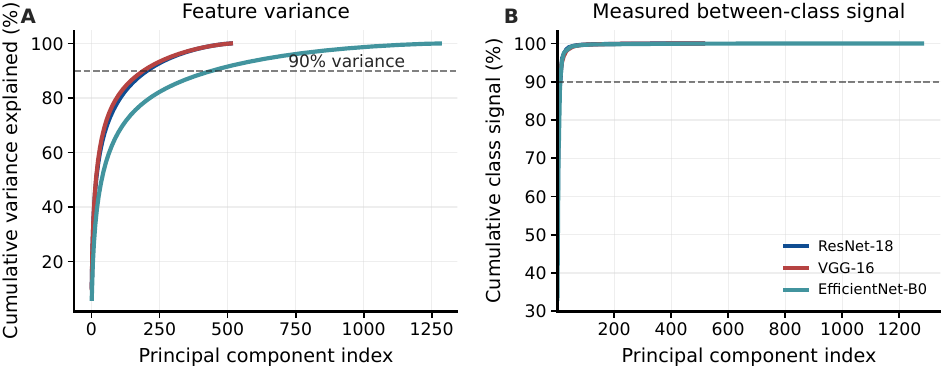}
\caption{\textbf{Pretrained feature spectra (CIFAR-10, seed 0).} (A) Cumulative feature variance. (B) Cumulative measured between-class signal $s_k$ from clean training labels; the dashed line marks 90\%.}
\label{fig:spectra}
\end{figure}

\Cref{fig:spectra} measures both premises of the rank argument. The first 60 components capture 70.8\%, 73.2\%, and 58.6\% of variance for ResNet-18, VGG-16, and EfficientNet-B0, respectively; their 90\% thresholds are ranks 205, 191, and 443. Class signal is much more concentrated: the first 40 PCs capture 98.9\%, 98.5\%, and 98.7\% of the measured $\sum_k s_k$, and 90\% is reached by ranks 11, 9, and 11. Thus later PCs add little centroid-level class separation in this dataset, while the rank sweeps below test whether exposing them helps fit corrupted labels. This directly measures $s_k$ but does not claim that the heuristic inequality in \cref{eq:gain} predicts the optimum.

\subsection{Main Result: Rank Sweeps Reveal Spectral Overfitting}

\Cref{fig:fixed_probe_main} and \cref{tab:cross_backbone} summarize the main finding. Across all three architectures: (i) clean labels favor full rank; (ii) noisy labels favor intermediate PCA ranks; (iii) random projections help but consistently less than PCA. We interpret the rank sweep as evidence that noisy probes overfit lower-variance directions of the frozen representation. In this CIFAR-10 setup, the effect is larger at higher noise rates and for EfficientNet-B0.

\begin{table}[!b]
\centering
\caption{\textbf{Cross-backbone fixed-probe results on CIFAR-10.} Mean $\pm$ std over seeds $\{0,1,2\}$. The best rank is selected independently per seed; the displayed rank is the modal value (``$\approx$'' marks variation), so rounded mean gains need not equal the difference of rounded accuracies. Frozen features, 5K train, $C=1$.}
\label{tab:cross_backbone}
\resizebox{\textwidth}{!}{%
\begin{tabular}{lcccccc}
\toprule
Backbone & Noise & Full-rank & Best PCA rank & Best PCA & PCA gain & Random gain \\
\midrule
R18 & 0\% & $82.6\pm0.2$ & 512 & $82.6\pm0.2$ & $+0.1\pm0.1$ & $+0.0\pm0.0$ \\
R18 & 20\% & $73.6\pm0.2$ & $\approx$60 & $79.0\pm0.3$ & $+5.5\pm0.2$ & $+2.4\pm0.2$ \\
R18 & 40\% & $64.0\pm0.8$ & $\approx$40 & $76.7\pm0.3$ & $+12.7\pm1.1$ & $+7.1\pm0.7$ \\
\midrule
V16 & 0\% & $80.7\pm0.3$ & $\approx$400 & $80.7\pm0.3$ & $+0.0\pm0.0$ & $+0.0\pm0.0$ \\
V16 & 20\% & $70.6\pm0.4$ & $\approx$100 & $76.6\pm0.2$ & $+5.9\pm0.2$ & $+2.4\pm0.6$ \\
V16 & 40\% & $60.4\pm0.4$ & $\approx$60 & $73.2\pm0.7$ & $+12.8\pm0.5$ & $+6.7\pm0.7$ \\
\midrule
EB0 & 0\% & $88.8\pm0.2$ & 1280 & $88.8\pm0.2$ & $+0.0\pm0.1$ & $+0.0\pm0.0$ \\
EB0 & 20\% & $68.3\pm0.8$ & $\approx$60 & $86.1\pm0.2$ & $+17.8\pm0.8$ & $+11.9\pm1.1$ \\
EB0 & 40\% & $49.0\pm0.6$ & $\approx$60 & $85.0\pm0.3$ & $+36.0\pm0.8$ & $+27.0\pm0.5$ \\
\bottomrule
\end{tabular}
}
\end{table}

\begin{figure}[t]
	\centering
	\includegraphics[width=0.88\linewidth]{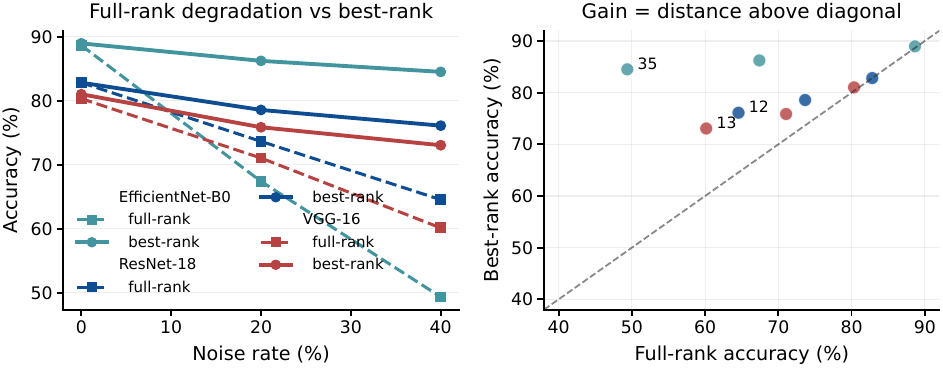}
\caption{\textbf{Full-rank degradation and best-rank resilience.} Left: full-rank degrades sharply with noise; best-rank preserves accuracy. Right: every noisy setting lies above the diagonal.}
	\label{fig:degradation}
\end{figure}

The largest case is EfficientNet-B0 at 40\% noise: full-rank drops to $49.0\pm0.6$\%, while the best PCA rank reaches $85.0\pm0.3$\%---within 3.8 points of the clean-label ceiling ($88.8\pm0.2$\%). Because the backbone is frozen, changing $d$ only changes the readout. \Cref{fig:degradation} shows that every noisy setting lies above the diagonal: the best rank outperforms the default full-rank probe.

\subsection{Diagnostics: Capacity, Spectra, and the Overfitting Gap}

The random projection baseline (\cref{fig:fixed_probe_main}B) separates rank reduction from PCA ordering. Across all backbones, random orthogonal projections provide roughly 40--75\% of the PCA gain. Thus part of the improvement is generic capacity reduction, while the remaining gap shows that PCA's ordering is useful in this setting. Together with the measured $s_k$, this is consistent with leading PCs preserving class signal while truncation limits noisy readout capacity; it is not a causal identification of where individual corrupted labels are stored.

\begin{figure}[t]
	\centering
	\includegraphics[width=0.90\linewidth]{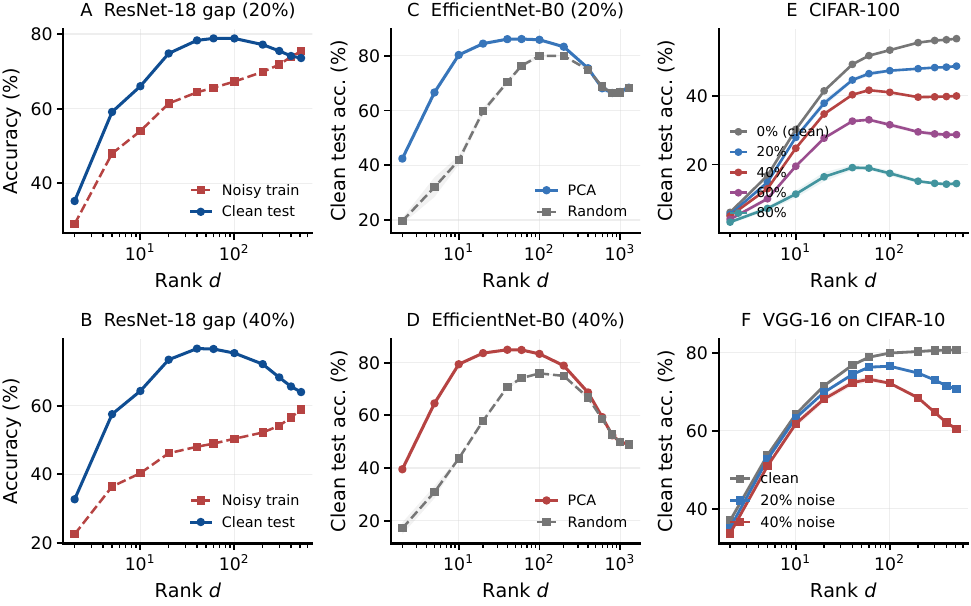}
	\caption{\textbf{Rank-sweep diagnostics and boundary cases.} (A--B) ResNet-18 train--test gap on CIFAR-10. (C--D) EfficientNet-B0 PCA vs.~random projections. (E) CIFAR-100 requires stronger noise for a clear elbow. (F) VGG-16 matches the CIFAR-10 pattern. All panels use rank $d$ on the horizontal axis.}
	\label{fig:support_grid}
\end{figure}

\begin{table}[t]
\centering
\caption{\textbf{Deployable rank selection and tuned-probe controls (CIFAR-10, 3 seeds).} 4K fit + 1K clean validation split. Values are mean $\pm$ std clean test accuracy.}
\label{tab:validation_controls}
\resizebox{\textwidth}{!}{%
\begin{tabular}{lccccc}
\toprule
Backbone & Noise & Full-rank default & PCA rank & PCA rank + tuned & Tuned full-rank \\
\midrule
R18 & 20\% & $72.2\pm0.6$ & $78.3\pm0.3$ & $79.2\pm0.1$ & $80.6\pm0.3$ \\
R18 & 40\% & $61.4\pm0.6$ & $75.9\pm0.9$ & $75.9\pm0.7$ & $78.6\pm0.8$ \\
V16 & 20\% & $69.1\pm0.3$ & $75.5\pm0.1$ & $77.2\pm0.2$ & $77.9\pm0.1$ \\
V16 & 40\% & $58.2\pm0.4$ & $71.7\pm1.0$ & $73.4\pm0.1$ & $75.6\pm0.4$ \\
EB0 & 20\% & $69.9\pm0.2$ & $85.4\pm0.2$ & $86.0\pm0.4$ & $87.4\pm0.3$ \\
EB0 & 40\% & $50.2\pm1.2$ & $84.1\pm0.8$ & $85.3\pm0.2$ & $85.9\pm0.2$ \\
\bottomrule
\end{tabular}
}
\end{table}

\Cref{tab:validation_controls} checks deployable model selection rather than oracle rank choice. The tuned full-rank probe is best in all six rows, exceeding the jointly tuned PCA probe by 0.6--2.7 points. This reversal is important: PCA rank selection is not the strongest deployment method here. Its value is diagnostic---the ordered rank path localizes where the default noisy readout begins to degrade---whereas ordinary $L_2$ tuning is the preferred baseline when clean validation data are available. Jointly tuning the selected PCA probe adds little.

\Cref{fig:support_grid}A--B provides an overfitting diagnostic. At 40\% noise with ResNet-18, the train--test accuracy gap at full rank is 21.7\,pt; the best diagnostic rank (rank 40) reduces it to 4.1\,pt. PCA truncation removes directions that improve noisy-train accuracy more than clean-test accuracy. \Cref{fig:support_grid}C--D shows that EfficientNet-B0 follows the same qualitative pattern at larger magnitude. The best PCA rank is $d=60$ for both 20\% and 40\% noise in our sweep, while the penalty for using the remaining dimensions grows with the noise rate.

\subsection{Boundaries: CIFAR-100, VGG-16, and Fine-Tuning}
\label{sec:boundaries}

CIFAR-100 is a harder transfer task, and the rank-sweep elbow is weaker at moderate noise (\cref{fig:support_grid}E). At 20\% noise, full rank remains optimal. At 40\% noise a small elbow appears (peak at $d=60$, $+1.7\pm0.2$\,pt). Increasing noise makes the elbow more pronounced: at 60\%, the gain is $+4.3\pm0.8$\,pt (33.0\% vs.\ 28.7\%); at 80\%, $+4.6\pm0.2$\,pt (19.0\% vs.\ 14.5\%). These small three-seed gains are boundary evidence, not a broad robustness claim. VGG-16 provides a second $D=512$ backbone (\cref{fig:support_grid}F): at 40\% noise, the best PCA rank reaches $73.2\pm0.7$\% vs.~$60.4\pm0.4$\% full-rank.

Finally, in one noisy fine-tuning control (ResNet-18, CIFAR-100, 20\% noise, trainable layer4 + classifier), PCA-truncated probes and classifier-head reconstruction favored high or full rank. From-scratch CIFAR controls behaved similarly. This single recipe shows that the fixed-feature result does not automatically transfer to learned representations; it does not establish a general fine-tuning conclusion.

\section{Conclusion and Limitations}
\label{sec:conclusion}

PCA rank sweeps reveal that default full-rank linear probes can overfit noisy labels along lower-variance directions of frozen pretrained representations. Across three supervised ImageNet CNNs on CIFAR-10, intermediate ranks gain up to $36.0\pm0.8$ points at 40\% noise versus the default $C{=}1$ probe; direct measurements show that the first 40 PCs retain at least 98.5\% of centroid-level class signal. However, tuned full-rank probes outperform validation-selected PCA probes in every tested setting, so the sweep is a diagnostic rather than a competitive noisy-label method. Evidence is limited to 5K CIFAR subsets, symmetric synthetic noise, three seeds, and CNN features; the small CIFAR-100 effect appears only at stronger noise ($+4.3\pm0.8$\,pt at 60\%), and one fine-tuning recipe is insufficient for generalization. Future work should test modern self-supervised and transformer representations, varying $N/D$, natural-noise benchmarks, early stopping, and noise-robust baselines.

\iftightsubmission
\begingroup
\footnotesize
\fi
\bibliographystyle{splncs04unsrt}
\bibliography{example_paper}
\iftightsubmission
\endgroup
\fi

\end{document}